\documentclass[conference]{IEEEtran}
\IEEEoverridecommandlockouts
\usepackage{cite}
\usepackage{amsmath,amssymb,amsfonts}
\usepackage{graphicx}
\usepackage{textcomp}
\usepackage{xcolor}
\usepackage{url}
\usepackage{multirow}
\usepackage{caption}
\usepackage{subcaption}
\usepackage{tabularx, booktabs}

\usepackage[ruled,vlined,linesnumbered]{algorithm2e}
\def\BibTeX{{\rm B\kern-.05em{\sc i\kern-.025em b}\kern-.08em
    T\kern-.1667em\lower.7ex\hbox{E}\kern-.125emX}}
\usepackage[colorlinks=true]{hyperref}

\newcommand{\dist}{{\rm dist}}

\begin{document}

\title{Accelerating Reinforcement Learning for Reaching using Continuous Curriculum Learning  \\
\thanks{}
}

\author{\IEEEauthorblockN{Sha Luo, Hamidreza Kasaei, Lambert Schomaker}
\IEEEauthorblockA{\textit{Bernoulli Institute}\\
University of Groningen, The Netherlands \\
s.luo@rug.nl, hamidreza.kasaei@rug.nl, l.r.b.schomaker@rug.nl}
}

\maketitle
\begin{abstract}
Reinforcement learning has shown great promise in the training of robot behavior due to the sequential decision making characteristics. However, the required enormous amount of interactive and informative training data provides the major stumbling block for progress. In this study, we focus on accelerating reinforcement learning (RL) training and improving the performance of multi-goal reaching tasks. Specifically, we propose a precision-based continuous curriculum learning (PCCL) method in which the requirements are gradually adjusted during the training process, instead of fixing the parameter in a static schedule. To this end, we explore various continuous curriculum strategies for controlling a training process. This approach is tested using a Universal Robot 5e in both simulation and real-world multi-goal reach experiments. Experimental results support the hypothesis that a static training schedule is suboptimal, and using an appropriate decay function for curriculum learning provides superior results in a faster way. 
\end{abstract}
\section{Introduction}
\begin{figure}[t!]
	\centering
	\begin{subfigure}[t]{0.2\textwidth}
		\centering
		\includegraphics[width=\textwidth]{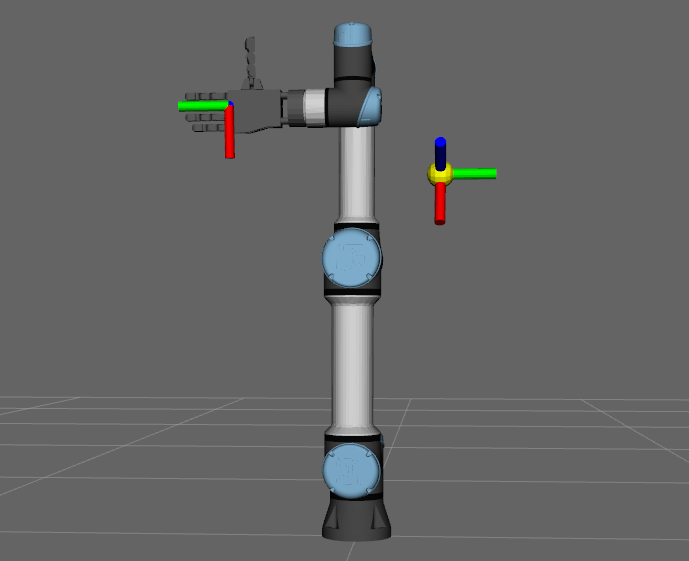}
		\caption[Initial pose]%
		{{\small Initial pose}}    
		\label{fig:process14}
	\end{subfigure}
	\quad
	\begin{subfigure}[t]{0.2\textwidth}  
		\centering 
		\includegraphics[width=\textwidth]{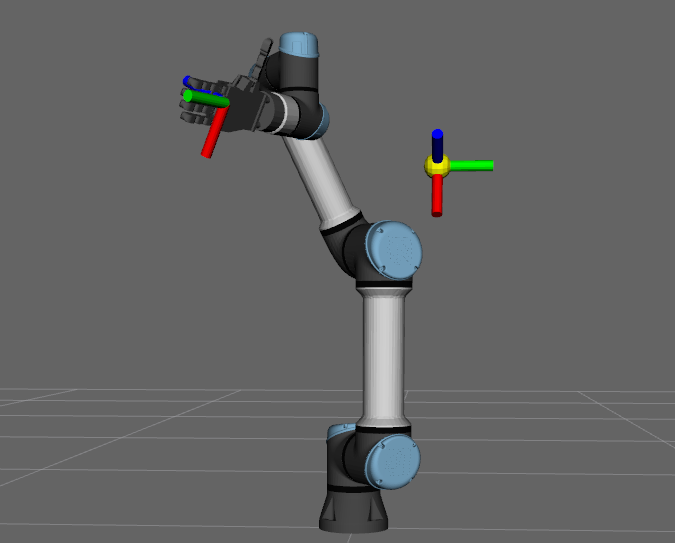}
		\caption[First step]%
		{{\small First step}}    
		\label{fig:process24}
	\end{subfigure}
	\vskip\baselineskip
	\begin{subfigure}[t]{0.2\textwidth}   
		\centering 
		\includegraphics[width=\textwidth]{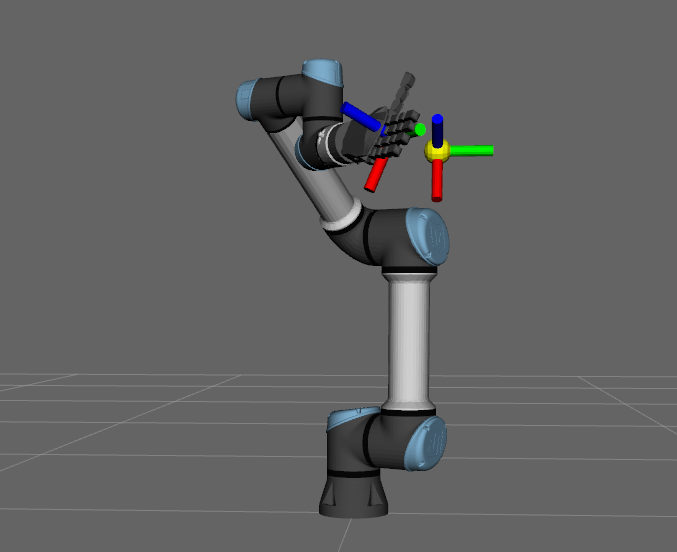}
		\caption[Second step]%
		{{\small Second step}}    
		\label{fig:process34}
	\end{subfigure}
	\quad
	\begin{subfigure}[t]{0.2\textwidth}   
		\centering 
		\includegraphics[width=\textwidth]{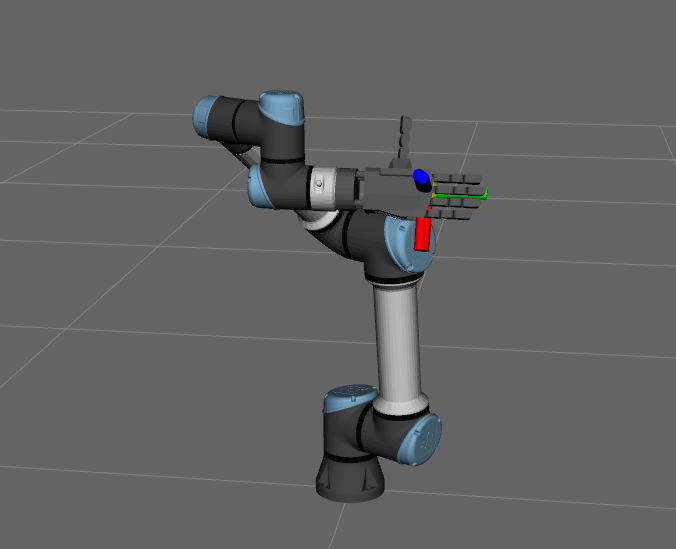}
		\caption[Final step]%
		{{\small Final step}}    
		\label{fig:process44}
	\end{subfigure}
	\caption[ Illustration of pose reach for grasping process]
	{\small Three steps of the reaching task for UR5e and qbhand. The yellow ball with an attached coordinate frame represents the target pose, while the reference frame on the palm represents the pose of the hand. (a) The initial pose of the arm. (b-c) Robot postures after the first and second steps bringing the robot's hand closer to the target pose. (d) The final success step: when the distance of the positions and orientations between the hand and target are within the required precision, the step is regarded as a successful move. The target pose is computed using forward kinematics based on randomly selected joint angles within stipulated joint limits. The number of steps required for a pose reach may vary in different target poses. 
} 
	\label{fig:process}
\end{figure}

In recent years, reinforcement learning (RL) has attracted growing interest in decision making domains, and achieved notable success in robotic control policy learning tasks \cite{gu2017deep} \cite{levine2016end} \cite{tai2017virtual} \cite{amarjyoti2017deep}. Reasons for this include RL requiring minimal engineering expertise to build complex robot models, and its framework for learning from interaction with the environment is suitable for training on sequential tasks. However, a low training efficiency is the primary challenge preventing reinforcement learning from being widely adopted in real robots. RL algorithms rely on experience that contains informative rewards, but random exploration methods usually only provide sparse rewards. Thus, the agent needs to acquire more experience to extract sufficient amounts of useful data, which results in long training time for a target policy. It is not practicable to train a robot for a long time due to safety factors. Then, in this study, we focus on improving the training efficiency of reinforcement learning for multi-goal reaching tasks\footnote{The reaching task in this study is defined as controlling the end effector to reach a pose: the positions in Cartesian space and the orientations in Euler angles.} in which the target is randomly generated in the workspace; an example of the reaching process is illustrated in Fig. \ref{fig:process}.

We are motivated by the concept of successive approximation, introduced by Skinner \cite{skinner1958reinforcement}\cite{peterson2000discovery}. He proposed this theory based on training a pigeon to bowl. This appeared to be a very tedious task because the desired behavior is far from the natural behavior of pigeons. Instead of rewarding the target behavior directly, they found that by positively reinforcing any behavior that is approximately in the direction of the desired final goal, the pigeon became a champion squash player within few minutes. From this example, we see that learning in animals is facilitated by appropriate guidance. They do not learn only from trial and error. The learning process is much faster through a suitable successive reward schedule with diverse guidance. The learned skills in easy early tasks can help them perform better in the later, more difficult situations. This concept has resurfaced in machine learning: curriculum learning (CL) \cite{bengio2009curriculum}, where the main idea is to train an agent to perform a defined sequence of basic tasks referred to as curriculum, and to increase the performance of agents and speed up the learning process for the final task. 

For reaching tasks in robots, different levels of precision requirements can be seen as a curriculum, making the problem more or less difficult as designed. Thus, the scientific research question is how to design the curriculum to enable efficiency, optimally. A poorly designed curriculum can also drive the learning process in the wrong direction, which should be avoided. In this study, we consider the required precision as a continuous parameter coupled with the learning process and propose precision-based continuous curriculum learning (PCCL) to improve the training efficiency and performance of multi-goal reach by a UR5e robot arm. At the beginning of the learning process, the reach accuracy requirement is loose to allow the robot to gain more rewards for acquiring basic skills required to realize the \textit{ultimate target}. To this end, the precision-based process is formulated with a cooling function that automatically selects the accuracy ($\epsilon$) based on the current stage of training.

The main contributions of this study are as follows: First, we propose a PCCL method for accelerating the RL training process. The experimental results showing that the proposed approach accelerates the training process and improves performance in both sparse reward and dense reward scenarios of multi-goal reach tasks. Secondly, the proposed PCCL is easily implemented in off-policy reinforcement learning algorithms. A similar average success rate in real world experiments demonstrate the feasibility of our method for physical applications without requiring fine-tuning when going from simulation to real-world applications. Finally, the robot's reaching movements in grasp and push tasks show that the learned reach policy is beneficial for advanced real robotic manipulations, which can pave the way for complex RL-based robotic manipulations. 

\section{Related Work}
The main goal of this study is to improve the training efficiency of reinforcement learning using curriculum learning. Although an extensive survey of efficient approaches for reinforcement learning in robotic control is beyond the scope of this study, we will review a few recent efforts.

Recently, various research groups have made substantial progress towards the development of continuous reinforcement learning for robot control. The long training durations for reinforcement learning algorithms demonstrate that it is difficult to train an agent to behave as expected using pure trial and error mechanisms \cite{plappert2018multi}\cite{andrychowicz2017hindsight}\cite{jurgenson2019harnessing}\cite{mahmood2018benchmarking}. Much effort has been invested on the exploration and exploitation issue. One of the conventional methods for accelerating the learning of a control policy is learning from demonstration: Ashvin et al. \cite{nair2018overcoming} coupled deep deterministic policy gradients (DDPG) with demonstrations and showcased a method that can reduce the required training time for convergence by giving nonzero rewards in the early training stages. Another popular technique is prioritized experience replay \cite{schaul2015prioritized}: shaping the experience by sampling more useful data. In this way, the exploitation ability to get more informative data for training is improved. Nachum et al. \cite{nachum2018data} showed how hierarchical reinforcement learning (HRL) can train the robot to push and reach using a few million samples and get the same performance as a robot trained for several days. 

For methods based on curriculum learning, hindsight experience replay (HER) \cite{andrychowicz2017hindsight} is one of the popular methods and can be considered a form of implicit curriculum that improves learning efficiency through learning from fatal experiences. It allows for skipping complicated reward engineering.  Instead of judging a behavior to be a fatal miss, it is classified as a successful attempt to reach another target. This approach of learning from a fatal experience can be viewed as a curriculum that trains the easier-met goals first. Kerzel et al. \cite{kerzel2018accelerating} accelerated reach-for-grasp learning by dynamically adjusting the target size. When the robot has little knowledge of the task, the Kerzel et al. approach increases the target size to make it easier to succeed and get positive rewards for learning the policy. Fournier et al. \cite{fournier2018accuracy} proposed a competence progress based curriculum learning method for automatically choosing the accuracy for a two-joint arm reach task.

The approaches used in the Kerzel et al. \cite{kerzel2018accelerating} and Fournier et al. \cite{fournier2018accuracy} studies have similarities with the approach proposed in this study, including employing a \textit{start-from-simple} strategy to accelerate the learning process, which can be categorized as curriculum learning. However, they use two-joint planar arm for a position reach environment, which is difficult to adjust into a more complicated environment. Furthermore, Kerzel's task simplification method deals only with discrete reach positions. Fournier's approach requires massive computation resources for multiple accuracy competence progress evaluations. Furthermore, it uses a discrete CL, and switching \textit{curriculums} between pre-defined accuracies poses the risk of instability when training for a large dimensional task.

The PCCL method we propose is based on the continuous curriculum learning. By changing the precision requirement ($ \epsilon $) up to every epoch, we add a new task to the \textit{curriculum}, which can smooth the training process and is easy to implement. Compared to the similar methods show in the Kerzel et al. \cite{kerzel2018accelerating} and Fournier et al. \cite{fournier2018accuracy} studies, our continuous curriculum strategy obviates intensive computation for learning status evaluation and can be utilized in a real robotic arm. 

\section{Background and Methodology}
Our PCCL algorithm is based on the continuous reinforcement learning mechanism, using the DDPG as the basic framework. Hence, we first introduce the background of RL and DDPG, followed by a description of reaching tasks. Then, the overall PCCL approach is illustrated.
\subsection{Reinforcement Learning}
We formalize our RL-based reaching problem as a Markov decision process (MDP). This process can be modeled as a tuple $(S, A, P, R,\gamma)$, where $S$ and $A$ denote the finite state and action space respectively, $P$ is the transition probability matrix $P:=P^{a}_{ss'}=\mathbb{P}[s_{t+1}=s'|s_{t}=s, a_{t}=a]$ representing the probability distribution to reach a state, given the current state and taking a specific action, $R$ is the state-action associated reward space and can be represented as $r_t$, given time $t$, and $\gamma \in [0, 1]$ is the discounted factor. The goal of the robot is to learn a policy $\pi$ by interacting with the environment. The deterministic policy can be represented as $\pi(a_{t}|s_{t})$, which guides the agent to choose action $a_{t}$ under state $s_{t}$ at a given time step $t$ to maximize the future $\gamma$-discounted expected return $R_{t}=\mathbb{E}[\sum_{i=0}^\infty\gamma^{i}r_{t+i+1}]$.
\subsection{Deep Deterministic Policy Gradients (DDPG)}
Because of the consideration of continuous state and action values, we restrict our interest to continuous control. Popular continuous control RL algorithms are: Deep Q-Network (DQN) \cite{mnih2015human}, Continuous Actor Critic Learning Automation (CACLA) algorithm \cite{van2007reinforcement}, and the Trust Region Policy Optimization (TRPO) \cite{schulman2015trust}, Proximal Policy Optimization (PPO) \cite{schulman2017proximal} and Deep Deterministic Deep Policy Gradient (DDPG) algorithms \cite{lillicrap2015continuous}. We selected DDPG as our algorithm platform, which is an RL algorithm based on actor-critic architecture. The actor is used to learn the action policy and a critic is applied to evaluate how good the policy is. The learning process involves improving the evaluation performance of the critic and updating the policy following the direction of getting higher state-action values. This algorithm can be seen as a combination of Q-learning and actor-critic learning because it uses neural networks as the function approximator for both actor and critic $(\theta^{Q}, \theta^{\pi})$. To solve the instability problems of two networks and the data correlation in experiences, it uses a replay buffer and target networks $(\theta^{Q'}, \theta^{\pi'})$.

The differentiable loss for the critic is based on the state-action value function:
\begin{equation}\label{loss critic}
Loss(critic) = Q(s_{t},a_{t}|\theta^{Q})-y_{t},
\end{equation}
where Q is the output of the critic network that satisfies the Bellman equation, yielding the expected state-action value; and $y_{t}$ is the real value:
\begin{equation}\label{real}
y_{t} = r_t+\gamma*Q(s_{t+1},a_{t+1}|\theta^{\pi'}).
\end{equation}
During the training process, the parameters of these two networks can be updated as follows:
\begin{equation}\label{update critic}
\theta^{Q}\leftarrow\theta^{Q}-\mu_{Q}\cdot\nabla_{\theta^{Q}}Loss(critic),
\end{equation}
\begin{equation}\label{update actor}
\theta^{\pi}\leftarrow\theta^{\pi}-\mu_{\pi}\cdot\nabla_{a}Q(s_{t},\pi(s_t|\theta^{\pi})|\theta^{Q})\cdot\nabla_{\theta^{\pi}}\pi(s_t|\theta^{\pi}),
\end{equation}
where the symbol $\nabla$ denotes gradients, $\mu_{Q}$ and $\mu_{\pi}$ represent the learning rate of these two networks, respectively. After the update of the actor and critic, the target networks' weights are merely the copy of the actor and critic networks' weights.

There are several reasons behind the choice of this algorithm: First, it works in both the continuous state and the action space, which eliminates the need for the action discretization in the use of DQN (the previously dominant method for continuous state space). Secondly, it is a deterministic algorithm, which is beneficial for robotics domains as the learned policy can be easily verified with reproduction once the policy has been converged, compared with stochastic algorithms such as TRPO and PPO. Thirdly, DDPG uses the experience replay buffer to improve stability by decorrelating transitions. This technique can be modified to use additional sources of experience, such as expert demonstrations \cite{jurgenson2019harnessing} \cite{nair2018overcoming}, which has been shown to be effective for improving the training process. 
\begin{figure}[t!]
	\centering
	\includegraphics[scale=1.0]{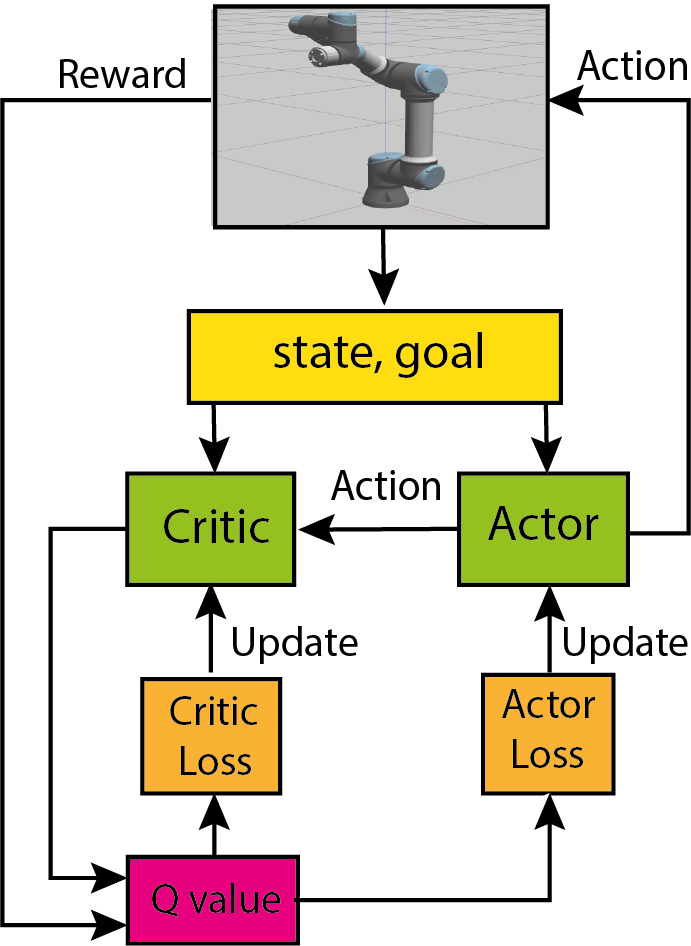}
	\caption{DDPG in a UVFAs framework}
	\label{fig:2}
\end{figure}
\begin{algorithm} 
	Initialize critic network $\hat{Q}(w,s,a,g|\theta^{\hat{Q}}) $ and actor network $\pi(w,s,g|\theta^{\pi})$ with weights $\theta^{\hat{Q}}$ and $\theta^{\pi}$ randomly\;
	Initialize target critic network $\hat{Q}'$ and actor network $\pi'$ weights
	$\theta^{\hat{Q}'} = \theta^{\hat{Q}}$ and
	$\theta^{\pi'} = \theta^{\pi}$\;
	Initialize experience replay buffer $R$\;
	\For{e in max\_epochs} 
	{
		Compute curriculum $\epsilon$ from equation \eqref{steepness}\;
		\For{i in max\_episodes}
		{
			$s, g = Env.reset()$ \;
			$g = g.append(\epsilon)$\;
			\For{j in max\_steps}
			{
				$a = epsilon\_greedy$ exploration\;
				$next\_s, r, done = Env.step(a, g)$\;	
				$Add (s, a, next\_s, r, g)$ into $R$\;
				$s = next\_s$\;
				\If{done}
				{
					break\;
				}
			}
		}
		\For{k in training\_steps}
		{
			Randomly sample N experience from $R$\;
			$y_{i} = r_{i} + \gamma Q'(s_{i+1}, g_{i}, \pi'(s_{i+1}, g_{i}|\theta^{\pi'})|\theta^{Q})$\;
			Critic and actor network weights update: following equations \eqref{loss critic}\eqref{real}\eqref{update critic}\eqref{update actor} \;
			Update the target networks:\;
			$\theta^{\hat{Q}'} = \tau \theta^{\hat{Q}}+(1-\tau)\theta^{\hat{Q}'}$\;
			$\theta^{\pi'} = \tau \theta^{\pi}+(1-\tau)\theta^{\pi'}$\;
		}
	}
	\caption{PCCL with DDPG based on UVFAs}
	\label{ddpg-algorithm}
\end{algorithm}
\subsection{Reach Task}\label{posereach}
The reach task is a fundamental component in robot manipulation that requires the end effector to reach to the target pose while satisfying Cartesian space constraints for various manipulation tasks. We aim to use reinforcement learning to train a UR5e on a multi-goal reach task, in which a value function (policy) is trained to learn how to map the current states into the robot's next action to reach the target goal. An example of a trained reach process is presented in Fig. \ref{fig:process}, where the UR5e starts from a fixed initial pose and follows the policy in taking actions to reach the goal pose.

A large body of recent studies has focused on position-only reaching \cite{pham2018optlayer} or scenarios where the end effector is forced to reach the targets from one direction, usually top vertical \cite{lenz2015deep}. Such simplification can greatly simplify the tasks but is difficult to generalize to interact with restricted objects. The skills learned for reaching are not easily used for further manipulations which normally require orientation constraints. We move the full joints of the arm for reaching one step forward, but only five joints are functioning. For the sake of simplicity, we couple the 4th joint angles with the 2nd and 3rd joints by always setting the 4th joint angles vertical to the ground. This is supported by the feature of the UR5e in that the 2nd, 3rd, and 4th joints are in the same plane. In this way, we also decrease the orientation dimension of the end effector from 3 to 2 as the $Roll$ axis is fixed. This joint coupling technique simplifies the task significantly while still being useful for right hand oriented reach tasks as it covers almost all of the reaching directions from the right side and the top.

In this study, we formulate the reach task as an MDP process with definitions of the $state$, $action$, $goal$ and $reward$:

\textbf{State:} The current pose of the end effector and joint angles, as defined in Eq. \eqref{state}. The pose is composed of positions and orientations of the end effector in which we set the position with meters as the unit and the orientation with the \emph{roll-pitch-yaw} order of the Euler angles in radian units. The angles of the $i$th-joint $j_{i}, i\in{\left\{1, \dots, 6\right\}}$ are also in radian units.
\begin{equation}\label{state}
s = [ee_{x}, ee_{y}, ee_{z}, ee_{Rx}, ee_{Ry}, ee_{Rz}, j_{i}],
\end{equation}

\textbf{Action:} We consider the joints' increments as the action (defined as $a$ in Eq. \eqref{action}),
which is a one-hot vector with the size of 6. The $a_{i}$ represents the normalized joint angle change for the $i$th joint. The angles of the $4$th joint is related to $a_{2}$ and $a_{3}$ to ensure its always vertical to the ground. 
\begin{equation}\label{action}
\begin{split}
a = [a_{i}];\quad i\in{\left\{1, \dots, 6\right\}}\quad a_{i}\in[-1, 1],\\
a_{4} =\left\{
\begin{aligned}
& -\pi-a_{2} & &  if\quad \left|\frac{\pi}{2}+a_{2}\right|+a_{3}\ge \pi,\\
& -a_{3}-a_{2}  & & \text{otherwise}.
\end{aligned}
\right.
\end{split}
\end{equation}

For stable training, we bound the action value within the range of -1 and 1. The maximum amount of change for the joint value is $ \pi/6 $. Thus, the action of -1 means decreasing the joint value by $ \pi/6 $, while 1 means increasing the joint value by $ \pi/6 $. Furthermore, the final computed actions are bounded by the joints limitations.
 
\textbf{Goal:} The goal is derived by uniformly sampling joint angles within the joint's limitations. Based on these joint angles, setting the forward kinematics result (end effector's pose) in Cartesian space as the goal can ensure a reachable target pose. 
\begin{equation}
goal = [ee_{x}, ee_{y}, ee_{z}, ee_{Rx}, ee_{Ry}, ee_{Rz}].
\end{equation}

\textbf{Reward:} Designing the reward function is a key to the success of RL. We compare our algorithm based on two types of reward: dense and sparse reward functions, as they are commonly used for reaching tasks. Thus, we give both definitions as follows: 

For the dense reward, the reward is given based on the Euclidean distance between the end effector and the goal pose. If both the position distance $\dist(p)$ and the orientation distance $\dist(o)$ are smaller than the required precision ($\epsilon$), we consider it a successful action and set the reward as 1. Otherwise, the reward is set to penalize the distance, which is a weighted summation of the position distance and the orientation distance.
\begin{equation}
	\dist(ee, goal) = \alpha\cdot\dist(p) + \beta\cdot\dist(o),
\end{equation}
where $\alpha, \beta \ge 0$ and $\alpha + \beta$ = 1. They are representing the weight factors of the position distance and orientation distance, respectively. In this study, these two distances are within the near range, and we set both weight factors to 0.5.
\begin{equation}
r =\left\{
\begin{aligned}
& - \dist(ee, goal)  & & {\epsilon < \dist(p)} \text{ and } {\epsilon < \dist(o)},\\
& 1 & & {\epsilon \ge \dist(ee, goal)}.
\end{aligned}
\right.
\end{equation}
For the sparse reward condition, a successful move is rewarded by 1, otherwise, 0.02 punishment is given for the energy consumption.
\begin{equation}
r =\left\{
\begin{aligned}
&-0.02  & & {\epsilon < \dist(ee, goal)}, \\
&1 & & {\epsilon \ge \dist(ee, goal)}.
\end{aligned}
\right.
\end{equation}

\subsection{Precision-based Continuous Curriculum Learning (PCCL)}
Curriculum learning is a technique that focuses on improving the training efficiency on difficult tasks by learning simple tasks first\cite{bengio2009curriculum}. Most of the previous studies in this field are focused on defining discrete sub-tasks as the \textit{curriculum} for agents to learn. For our multi-degree-of-freedom (DoF) reach environment, the shift between discrete precision can easily result in an unstable performance as the skills learned in the previous tasks also require time to adjust to the new, stricter tasks. However, the continuous CL can smooth the learning by continuously shrinking the precision. 

In this section, we focus on a precision-based continuous curriculum learning (PCCL) method, which uses a continuous function to change the curriculum (required precision). The scientific challenge arises how to design a function that works as the \textit{generator} of the curriculum, just like the job of a teacher to choose the practical level of curriculum for the students. 

\textbf{Framework Definition:}
A detailed mathematical description of the discrete CL framework has been formulated in the study by Narvekar et al. \cite{narvekar2016source}. Bassich et al. \cite{Bassich2019} extended it with decay functions as the curriculum generator for continuous CL. 
We formulate our continuous curriculum learning model based on their work. The task domain is described as $D$, which is associated with a vector of parameters $F^{t}=[F_{0}, ...F_{n}]$. The curriculum \textit{generator} is described as $\tau$, which is used to change the curriculum $M_{t}=\tau(D, F^{t})$ \cite{narvekar2016source}. An important rule when designing the \textit{generator} is that the difficulty should increase monotonically. We refer to the difficulty of the curriculum in the environment as $O$. Thus, it should satisfy the constraint: $O(M_{t})\leq O(M_{t-1})$.

Compared with discrete CL, the distinguishing feature of continuous CL is the parameter $t$. It can be referred to as an episode, which means changing the curriculum up to every episode. We refer to the parameter $t$ as the epoch number for updating the policy. 

For the decay function, Bassich et al. \cite{Bassich2019} categorized it into two classes: fixed decay and adaptive decay functions, based on if it depends on the performance of the agent when deriving the curriculum. The fixed decay functions have the benefit of being easily implemented with only a few parameters to be considered, with the downside being that it lacks flexibility. The adaptive decay functions may generalize to different configurations, but they take up much time evaluating the performance to get feedback for computing the suitable curriculum, especially for complex high-dimensional environments. Furthermore, designing the adaptive decay function requires prior expert knowledge. Thus, we choose the fixed decay function for generating the training curriculum. Once the parameters of the fixed decay function are well designed at the outset, the policy can enjoy the advantages continuously. 
We formulate the decay function based on a cooling schedule \cite{schomaker2004automatic} to generate the curriculum. It contains four parameters: start and end precision ($e_{0}$ and $e_{m}$ respectively), the number of epochs that a decay function should experience ($s$), and the monotonic reduction slope of the decay ($\alpha$). The decay function is formalized as a power function \eqref{steepness}:
\begin{equation}\label{steepness}
\begin{split}
\epsilon = e_{m}+\left(\frac{s-k}{s}\right)^{\alpha}(e_{0}-e_{m}),
\end{split}
\end{equation}
where $k\in{[0, s]}$ represents the current epoch number, $s$ is the total number of epochs for the entire precision reduction process, and $\alpha\in{(0, \infty)}$ is the slope of the decay function. The precision ($\epsilon$) of the training process decreases gradually following the equation \eqref{steepness}. If $\alpha$ is smaller than 1, the initial decaying process is slower with a smaller slope $\alpha$. The decay function is a linear function when $\alpha$ is equal to 1. If $\alpha$ is larger than 1, then the larger of the slope, the faster the decay of the initial part of the precision behaves. 

The implicit problem with using CL in the precision ($\epsilon$) for a reaching task is that it can violate the MDP structure of the reach problem by changing the requirements for a successful reach. In this way, the state can be terminated or under different precision requirements, leading to fake updates in the RL algorithm. One of the methods to solve this problem is using universal value function approximators (UVFAs), as proposed by Schaul et al. \cite{schaul2015universal}. It formalizes a value function to generalize both states and goals. In our multi-goals reach environment, the goals are generated randomly in the entire workspace, which means the training policy not only need to consider the state but also the goals. Furthermore, in different stages of the training process of our proposed PCCL method, the precision level for goals are different, which makes the model perfect to be used with the architecture of the UVFAs. Under the architecture of the UVFAs, the goal of the environment is composed of the pose of the targets, and the precision of a successful reach. Therefore, the inputs for the actor and critic in DDPG are being extended with the renewed goal, updating those fake experience to the right one that meets the MDP feature. The framework of the UVFAs based DDPG algorithm can be found in Fig. \ref{fig:2}. We also illustrated the pseudo-code of our proposed algorithm in Algorithm. \ref{ddpg-algorithm}. 
\section{Experiments}
In this section, we first introduce the environmental setup and baseline. Because there are no standard environments for multi-goal reach, we create our environment using the Gazebo simulator and setup a Python interface between the RL algorithm and the agent environment. We then compare the performance and efficiency of our proposed PCCL approach with that of the vanilla DDPG in the context of various reach tasks performed by a UR5e robot arm in both simulation and real world environments.    
\subsection{Environmental Setup}
All of the experiments are conducted in the Ubuntu 16.04 system and ROS 'Melodic' package. Gazebo 9.0 is used for simulation. The system is encapsulated into a Singularity container\footnote{\url{https://sylabs.io/guides/3.4/user-guide/}}, which can be easily reproduced with a bootstrap file in any environment. For the parameter search process, the university's HPC facility with GPU nodes was used.

On the algorithm side, a UVFAs based multi-goal DDPG is used as the baseline, in which the robot is told what to do by using an additional goal as input. The pipeline of the algorithm is illustrated in Fig. \ref{fig:2}. The input layer of the actor is composed of state and goal. The output for the actor is the joint angles increments for all of the six joints. For the critic, the input layer is composed of current state, action, and goal, while the output is the Q value for the actions given the state and goal. Both of the actor and critic networks are equipped with three hidden layers with the size of (512, 256, 64), followed by the same ReLU \cite{agarap2018deep} activation function. For the output layer, the critic network does not use an activation function to keep the real Q value (state value estimation) unbounded, while the policy network uses the $\tanh$ activation function. The DDPG hyper-parameters used in the experiments are listed in Table \ref{tab:ddpg-para}. For the hyper-parameters in PCCL, the start curriculum means the starting reach precision for the position and orientation. For the dense reward setting, we set the starting precision ($\epsilon$) to 0.15 (units for the position: m, units for the orientation: radian). For the sparse reward setting, it is harder to get positive rewards. Through the experiment, we find out that 0.25 is the start precision at which something can be learned using the sparse reward setting. We, therefore, enlarge the start precision to 0.25 for the sparse reward setting. Furthermore, because the sparse reward environment is sensitive to the change of the precision ($\epsilon$), we design the curriculum changing in a much slower way with 2.5K epochs compared to 1.0K with dense reward to reduce to the final precision. In both of the settings, a slope of 0.8 is selected for the decay function as it works well for both rewards settings.
\begin{table}[!htbp]
	\centering
	\caption{DDPG Training Hyperparameters}
	\begin{tabular}{ c c c }
		\hline
		\hline
		Hyperparameters&Symbol&Value\\ 
		Discount factor&$\gamma$&0.98\\
		Target update ratio&$\tau$&0.01\\
		Actor learning rate&$\mu_{Q}$&0.0001\\
		Critic learning rate&$\mu_{\pi}$&0.001\\
		Gaussian action noise&$\sigma$&0.1\\
		Replay buffer size&$B$&5e+6\\
		Batch size&$N$&128\\
		Epochs of training&$E$&3K\\
		Episodes per epoch&$M$&10\\
		Steps per episode&$T$&100\\
		Training steps per epoch&$K$&64\\
		Exploration method&*&$\epsilon\_greedy$\\

	\end{tabular}
	\label{tab:ddpg-para}
\end{table} 

\subsection{Simulation Experiments}
We evaluate the training efficiency and performance of the proposed approach against the vanilla DDPG algorithm on both sparse reward and distance-based dense reward scenarios.  

\begin{figure*}[t!]
	\centering
	\begin{subfigure}[t]{0.37\textwidth}
		\centering
		\includegraphics[width=\textwidth]{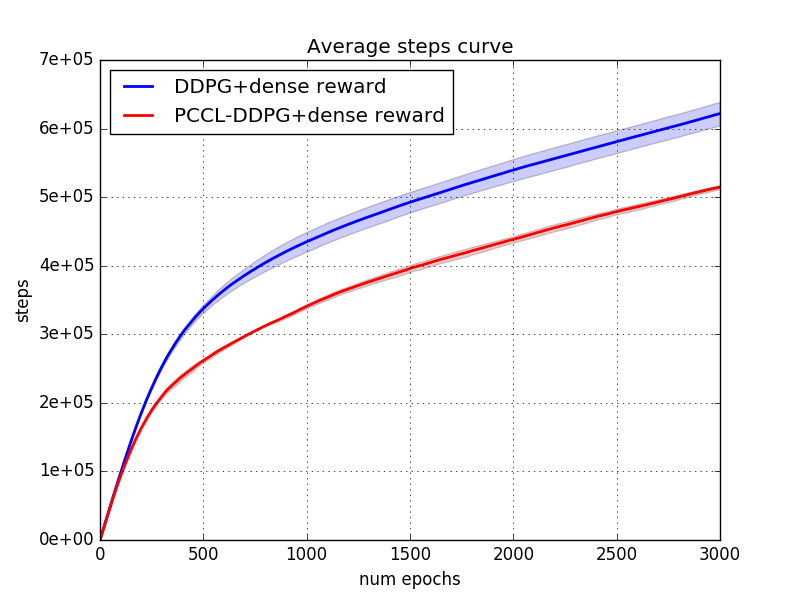}
		\caption[Initial pose]%
		{{\small Average accumulated steps with dense reward.}}    
		\label{fig:steps_dense}
	\end{subfigure}
	\quad
	\hspace{.3in}
	\begin{subfigure}[t]{0.37\textwidth}  
		\centering 
		\includegraphics[width=\textwidth]{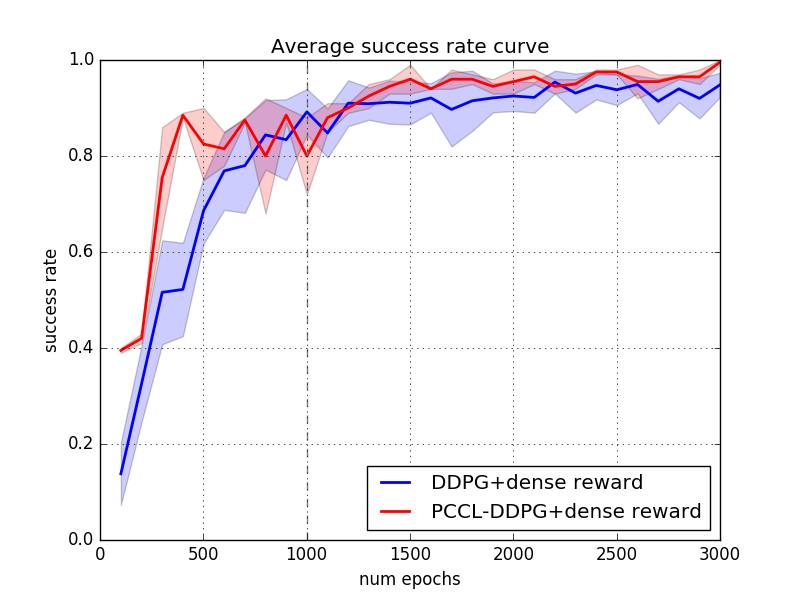}
		\caption[First step]%
		{{\small Average test success rate with dense reward.}}    
		\label{fig:success_dense}
	\end{subfigure}
	
	\begin{subfigure}[t]{0.37\textwidth}   
		\centering 
		\includegraphics[width=\textwidth]{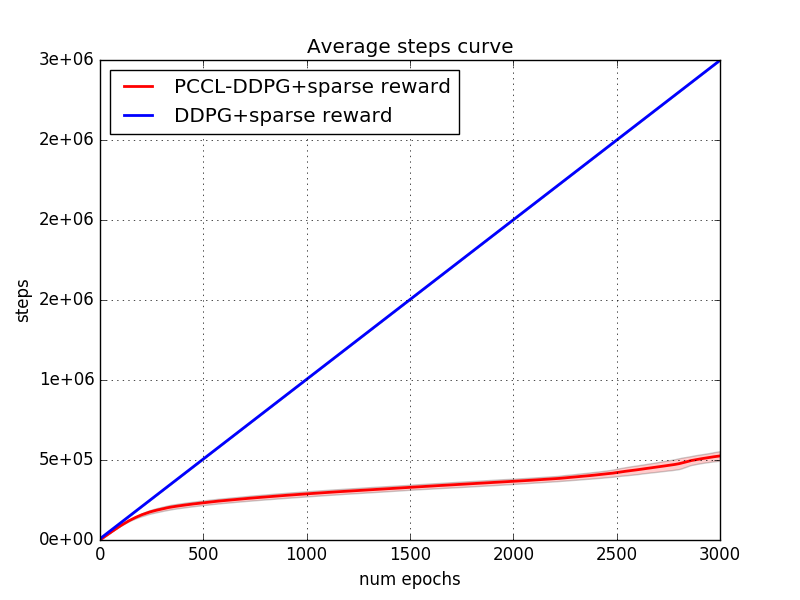}
		\caption[Second step]%
		{{\small Average accumulated steps with sparse reward.}}    
		\label{fig:steps_sparse}
	\end{subfigure}
	\quad
	\hspace{.3in}
	\begin{subfigure}[t]{0.37\textwidth}   
		\centering 
		\includegraphics[width=\textwidth]{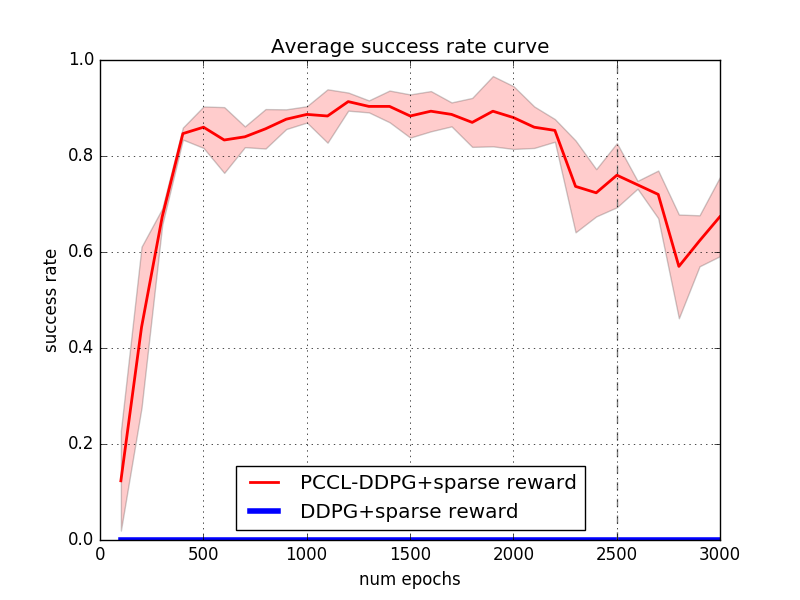}
		\caption[Final step]%
		{{\small Average test success rate with sparse reward.}}    
		\label{fig:success_sparse}
	\end{subfigure}
	\caption[ Pose reach for grasping process illustration]
	{\small Simulation results on average accumulated steps, i.e., accumulated simulation steps during the learning process, and average success rate: The shaded regions represent the variance over ten runs. The gray dash line indicates the epoch that arriving the final curriculum, after which the PCCL method has the same precision criteria as the baseline.} 
	\label{fig:sim-exp}
	\vspace{-3mm}
\end{figure*}

At the beginning of each episode in training, the robot is initialized to have the same joint angles: $[-\frac{\pi}{2}, -\frac{\pi}{2}, 0, \frac{\pi}{2}, \frac{\pi}{2}, -\frac{\pi}{2}]$, and then is trained to reach a target pose that is derived from a randomly selected joints configuration. If the pose distance is within the precision before running out of the maximum steps, this step is regarded as a successful reach and the episode is being terminated. We record the accumulated steps every 10 epochs and evaluate the averaged success rate every 100 epochs over 100 various target poses. The simulation experiment results can be seen in Fig. \ref{fig:sim-exp}. Our PCCL method is able to improve the performance in both rewards settings, especially for the sparse situations, where PCCL allows the task to be learned at all compared to the multi-goal DDPG baseline (from 0.0\% to 75\%). Besides, from Fig. \ref{fig:steps_dense}, we can see that our proposed approach can reduce the number of necessary steps by nearly $1.0e+05$ with less variance in the dense-reward situation. For reinforcement learning, an efficient policy could learn the skills faster and result in an earlier termination in episodes, thus requires fewer steps during the training period, implying less time on training. Specifically, the total training time is decreased by $19.9\%$ in the dense-reward condition, which is summarized in Table \ref{tab:time}. In the sparse-reward setting, the training time of our PCCL method is even less than the dense-reward setting with the normal multi-goal DDPG algorithm. 

Once the policy has been learned in simulation, we evaluate the average success rate in both simulated and physical UR5e robots over 100 runs. The averaged results are shown in Table \ref{tab:success_rate}. Based on the obtained results, it is obvious that the policy trained in simulation can be applied to the real robot without any further fine tuning. Furthermore, the reach policy learned in simulation can be tested in the real world with almost the same performance. 
\begin{figure*}[t!]
\vspace{-5mm}
	\centering
	\begin{subfigure}[t]{0.3\textwidth}
		\centering
		\includegraphics[width=\textwidth]{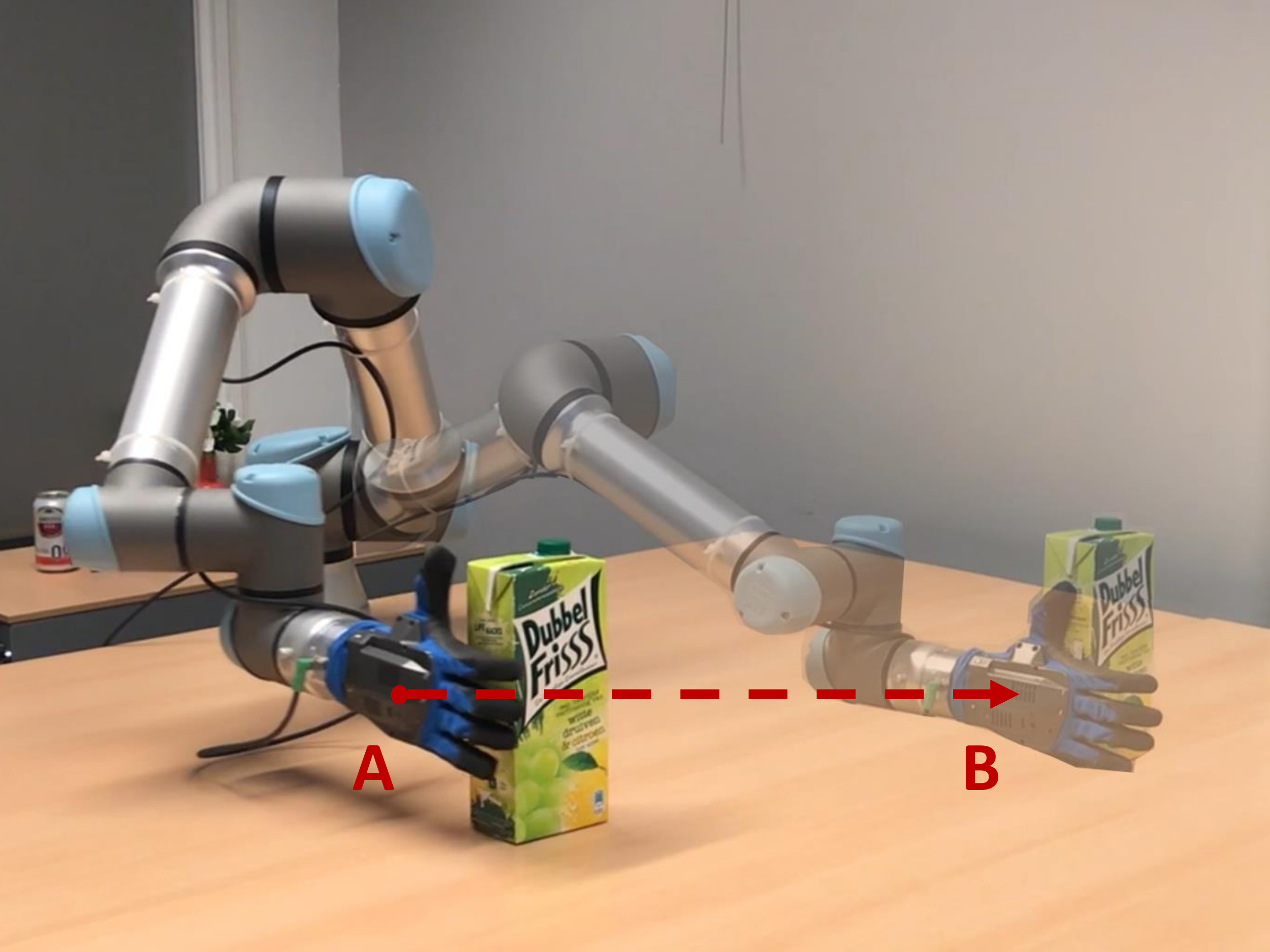}
		\caption[Initial pose]%
		{{\small Push scenario}}    
		\label{fig:push}
	\end{subfigure}
	\quad
	\hspace{.7in}
	\begin{subfigure}[t]{0.3\textwidth}  
		\centering 
		\includegraphics[width=\textwidth]{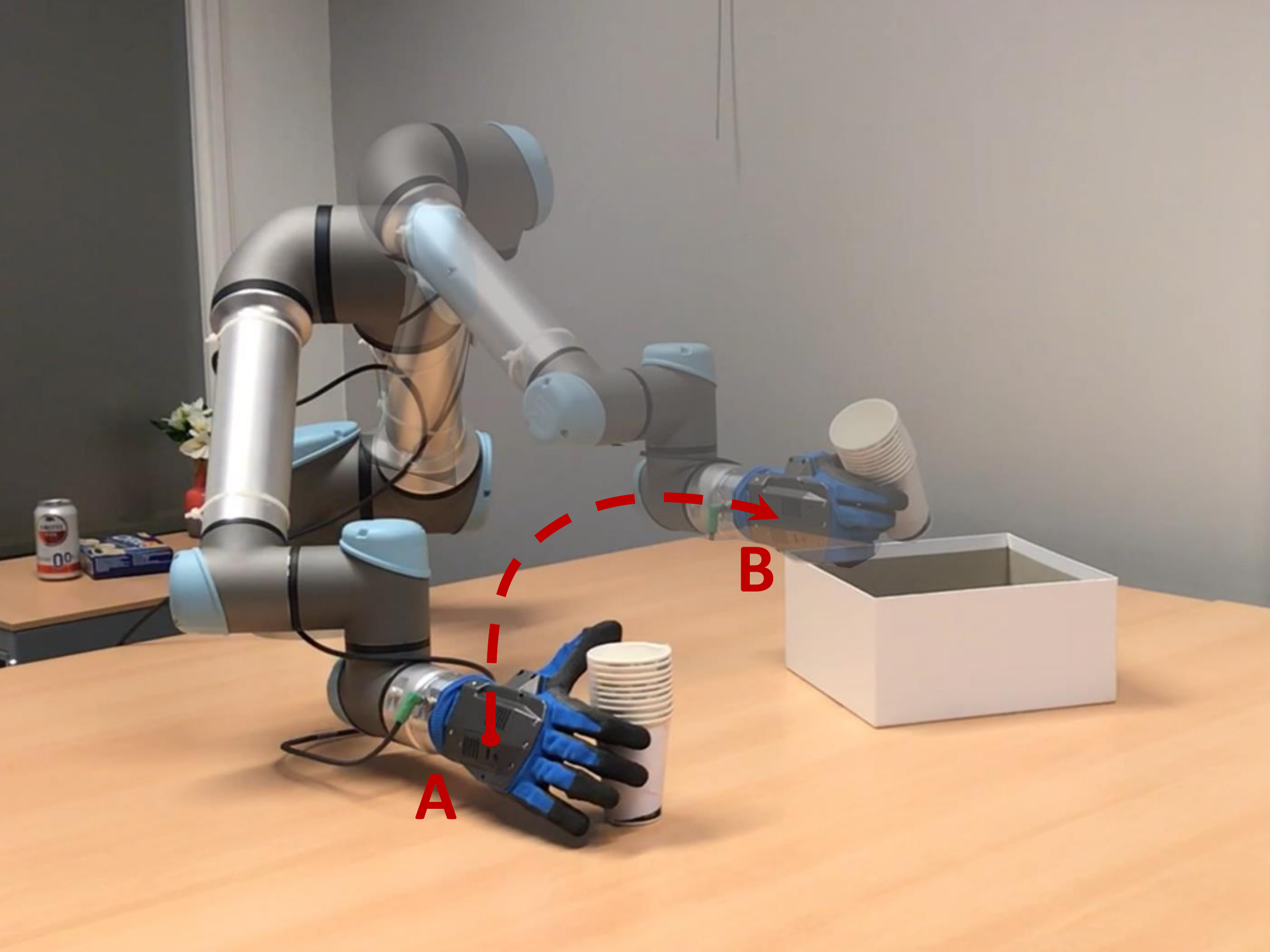}
		\caption[First step]%
		{{\small Grasp scenario}}    
		\label{fig:grasp}
	\end{subfigure}
	\caption[ Pose reach for grasping process illustration]
	{\small Push and grasp experiments: the proposed algorithm is mainly responsible for moving the robotic arm to the target pose. Object perception and object grasping are done by the RACE system \cite{kasaei2018towards}\cite{kasaei2019interactive}. In the case of push scenario (a), the arm first goes to the pre-push pose, as shown by a red point A. Then, the robot pushes the juice box object from point A to the target pose B. In the case of object grasping scenario (b), the robotic arm first reaches the pre-grasp pose as shown by the red point A, then grasps the cup and manipulates it on top of the basket, as shown by point B.} 
	\label{fig:manipulation}
	\vspace{-5mm}
\end{figure*}
\begin{table}[!htbp]
	\centering
	\caption{Training time comparison}
	\begin{tabular}{|c|c|c|c|c|c|c|}
		\hline
		\multirow{2}*{Methods}&\multicolumn{2}{|c|}{Total time}\\  
		\cline{2-3}
		~&Dense reward&Sparse reward\\
		\hline
		DDPG&22h29min6s&52h30min\\
		\hline
		PCCL-DDPG&18h01min33s&19h41min20s\\
		\hline
	\end{tabular}
	\label{tab:time}
\end{table} 

\begin{table}[!htbp]
	\centering
	\caption{Performance on trained policy}
	\begin{tabular}{|c|c|c|c|c|c|c|}
		\hline
		\multirow{3}*{Methods}&\multicolumn{4}{|c|}{Success rate (\%)}\\  
		\cline{2-5}
		~&\multicolumn{2}{|c|}{Dense reward}&\multicolumn{2}{|c|}{Sparse reward}\\
		\cline{2-5}
		~&Sim&Real&Sim&Real\\
		\hline
		DDPG&0.950&0.950&0.000&0.000\\
		\hline
		PCCL-DDPG&0.982&0.980&0.715&0.705\\
		\hline
	\end{tabular}
	\label{tab:success_rate}
\end{table} 

\subsection{Real World Experiments}
We test the manipulation performance on two tasks: pushing and grasping based on the reaching policy learned by our PCCL algorithm. In this experiment, a UR5e arm, fitted with a Qbhand, is used to test the performance of the proposed approach. Towards this goal, the proposed approach is integrated into the RACE robotic system \cite{kasaei2018towards} \cite{kasaei2019interactive}. In particular, the RACE framework provides the pose of active objects as the goal to reach by the arm. The experimental settings for pushing and grasping tasks are shown in Fig. \ref{fig:manipulation}.

For the pushing task, a juice box is randomly placed at a reachable position in the workspace. The initial goal is to move the arm to the pre-push area and then push the juice box from point A to point B as shown in Fig. \ref{fig:push}. The robot is commanded to reach different target poses to accomplish this task. 

For the grasping task, a cup and a basket are randomly put on the table that can be reached by the arm. The initial goal of this task is to move the arm to go to the pre-grasp pose of the cup (i.e., represented as point A in Fig. \ref{fig:grasp}) using the trained PCCL algorithm. Once it reaches the target pose, the RACE system will take over the responsibility of controlling the hand to grasp the cup. After grasping the object, the robot moves the cup to the top of the basket using the PCCL algorithm (i.e., reach to pose B, as shown in Fig. \ref{fig:grasp}). Finally, the robot releases the cup into the basket. A video of the experiment is available\footnote{see \url{https://youtu.be/WY-1EbYBSGo}}. These demonstrations show that although the policy is trained on the initial \textit{up} pose, it can be generalized to other different initial states. Furthermore, the policy learned in simulation can be directly applied on the physical robot. Additionally, the learned skill can be further utilized in more complicated scenarios, as shown by the success of the manipulation tasks. Accelerating the training process of RL for reaching tasks may benefit not only the RL field but the robotics autonomy domains as well. 

\section{Discussion and Conclusion}
We have proposed a precision-based continuous curriculum learning (PCCL) approach to accelerate reinforcement learning for multi-DOF reaching tasks. The technique utilizes the \textit{start-from-simple} rule to design a continuous curriculum \textit{generator}: decay function. The experimental results show that the proposed method can improve the RL algorithm training efficiency and performance, especially for a sparse and binary reward scenario, which fails to learn the task in the baseline environment. 

Compared to other curriculum learning based methods on multi-goal reach tasks \cite{kerzel2018accelerating}\cite{fournier2018accuracy}, the proposed approach enables the training of a high-dimensional pose reaching task by deploying a continuous-function based curriculum strategy. Although the predefined decay function lacks general flexibility compared with adaptive functions, it is an efficient way to learn complex tasks and environments. Conversely, the adaptive functions would be required to evaluate the difficulty level of the environment and trained policy status frequently during the learning process, which normally means success rate evaluation or environment analysis. Those evaluation steps entail a heavy computation burden for the agents and contain delayed information for training. Our method can be generalized to different robotic arms and environment. The main limitation is that the difficulty of the situation should be related to the precision requirements; for instance, the famous Fetch environment from OpenAI \cite{brockman2016openai}.

Note that although the reach task that we consider in this study is simple, it mainly serves as a convenient benchmark to straightforwardly demonstrate the significant improvement of training speed with our proposed PCCL method in high-dimensional continuous environment. By improving the efficiency of training RL algorithm, we, to some extend, pave the way for further investigation of some RL-based robotics applications, because the trained reach policy is a preliminary step towards further real world complex robotics manipulation tasks. As a future work, we plan to take obstacle avoidance into consideration, and explore the possibility of curriculum learning on complex motion planning. 

\section*{Acknowledgment}
We are thankful to Dr. Marco Wiering for his thoughtful suggestions. We also thank Weijia Yao for the fruitful discussion about the experimental design and the useful comments for the paper. Sha Luo is funded by the China Scholarship Council.

\bibliographystyle{IEEEtran}
\small{\bibliography{ref}}
\end{document}